# The Relativity of Induction

**Larry Muhlstein**
Department of Cognitive Science
University of California San Diego
San Diego, CA 94703
lmuhlstein@ucsd.edu

## Abstract

Lately there has been a lot of discussion about why deep learning algorithms perform better than we would theoretically suspect. To get insight into this question, it helps to improve our understanding of how learning works. We explore the core problem of generalization and show that long-accepted Occam's razor and parsimony principles are insufficient to ground learning. Instead, we derive and demonstrate a set of relativistic principles that yield clearer insight into the nature and dynamics of learning. We show that concepts of simplicity are fundamentally contingent, that all learning operates relative to an initial guess, and that generalization cannot be measured or strongly inferred, but that it can be expected given enough observation. Using these principles, we reconstruct our understanding in terms of distributed learning systems whose components inherit beliefs and update them. We then apply this perspective to elucidate the nature of some real world inductive processes including deep learning.

## Introduction

We have recently noticed a significant gap between our theoretical understanding of how machine learning works and the results of our experiments and systems. The field has experienced a boom in the scope of problems that can be tackled with machine learning methods, yet is lagging behind in deeper understanding of these tools. Deep networks often generalize successfully despite having more parameters than examples[1], and serious concerns[2] have been raised that our full suite of theoretical techniques including VC theory[3], Rademacher complexity[4], and uniform convergence[5] are unable to account for why[6]. This suggests that we do not understand generalization in quite the right way[2][7]. To build a better sense of the nature of learning, it helps to start with the fundamentals.

Induction is a process of extrapolating from data. We fit a pattern to our observations and use it to predict what we have not yet seen. But there are infinite patterns that fit any data. Some will successfully predict the future; others will not. The problem of induction boils down to this problem of choosing the best patterns/hypotheses/explanations amongst the sea of possibilities. We refer to this difficulty as the *underdetermination of induction*.

The philosophical tools for performing this selection–Occam's razor[8] or principles of parsimony[9]– state that we should choose the simplest pattern that fits as our primary hypothesis. The justification for this is often intuitive. Simplicity has been used to explain induction at least since Aristotle[10] and appears to have helped in a myriad of uses. However, attempts to make this notion precise have resulted in a plethora of formalisms and principles that only somewhat commute. In this essay, we explore the nature of these principles and demonstrate that there is no privileged notion of simplicity that can guide induction. Instead of relying on generalization techniques based on model complexity and data sample size, we show that a grounded explanation of the performance of inductive inference is achievable only by examining the holistic process by which our arbitrary prior beliefs come to reflect the structure of the world.

We argue that the assumptions used to guide most theoretical approaches to learning are applied based on empirical and imperfect beliefs about our systems of study, and therefore that they may only be used as heuristics. Without these assumptions, we cannot generally benefit from statistical guarantees that emerge from principles like independence. We demonstrate that general inductive inference is facilitated by a correspondence between a model and process, and that the only reliable way to achieve this correspondence is by proceeding from a state of unknown model performance to improved performance via extended causal chains of observation, belief update, and belief transfer. All learning occurs relative to a set of prior beliefs, and generalization performance may not be measured in absolute–only as improvement.

Using these results on the relativistic nature of induction, we argue that an understanding of the effectiveness of contemporary machine learning methods requires a broader view of how researchers' beliefs affect decisions about model structure, training process, and hyperparameter choice, as well as how these beliefs emerge from scientific experimentation and communication. By recognizing the role of prior learning in the formation of the inductive biases of our machine learning models, we can not only come to see why these models are so much more effective than current theory indicates, but how the inductive nature of the scientific process, human culture, and even biological evolution form a single learning system that serves to improve our understanding of our world via observation, communication, and revision of belief.

## A general construction

In order to explore the nature of induction, it is helpful to establish basic concepts. Learning involves some process we are trying to predict, some observations of data generated by that process, and other unobserved data that might be generated by the process in the future. There is also a learner or model whose goal is to observe data from the process and use these observations to improve at predicting future data generated by that process.

Often in machine learning theory, we study learning in constrained scenarios. For instance, we can prove strong guarantees on the sample complexity of learning a linearly-separable concept[11]. This is a viable approach because we have analytically[12] stipulated that the true concept is linearly separable. While this can be done when working with theory, it doesn't apply to most real world learning. In many learning scenarios, we may have a strong suspicion that a concept is roughly linearly-separable and therefore decide to use a linear SVM[13] to approximate it, but this is just a heuristic. Furthermore, this belief is fallible, as it emerges from its own inductive process where we observe data from the world and update our own beliefs. If we follow this rabbit hole, we will inevitably find that in any true synthetic process that we wish to predict, we cannot rule out any hypotheses or benefit from stronger guarantees than are possible in the general induction scenario. This is important because it means that, if we really want to understand the nature of induction processes, we need to study the fully general empirical problem,[1] even when the proximal process we are trying to predict appears to come largely from a simpler class.

To maximize generality, we do not assume that all data are generated by drawing at random from a probability distribution. Rather there may exist some arbitrary sequential relationship between the data. This allows us to capture things like causality, state, and other complex dependencies with prima facie unclear bounds. Formally, we can think about a piece of data as an element from a finite set of possible observations. We denote this set as the *alphabet* or $a$. The alphabet represents a part of the interface between the learner and the world in that it defines the kinds of information or distinctions that the learner can perceive. It is the same concept as in information theory[14].

As data are observed, we accumulate an ordered sequence of observations that can be represented via a finite series or *string* $o \in O$ of elements from a where the first element in the string is the first observation and so on.[2] We also have an infinite *sequence* $s_a$ of data that the process will generate, which represents the full extent of what one would observe if they were to make observations ad infinitum. In the context of a model, we may alternatively refer to sequences as *hypotheses* $h \in H$. For a given learner observing from a given process $p \in P$, there is exactly one sequence of data that will be observed.[3] We call this the *true hypothesis*.

---

[1] Some problems such as chess really are analytic, in which case this theory may not apply.

[2] In cases where choice of alphabet is clear or unimportant, we drop the subscript to streamline the notation.

[3] For simplicity of presentation, we assume determinism. This is without loss of generality.

By considering all possible sequences of observed data, we absolve ourselves from the need to enumerate and describe all of the possible complex-structured models. For instance, non-iid models may have a Markov property depending on an arbitrary number of steps back. There might not even be a limit to the length of the dependency. We don't want to have to worry about this.

If we naively attempt to assign a probability distribution to all possible sequences of data, we will find that the set is uncountable and that this cannot be done. To ameliorate this, we use a trick from algorithmic information theory and consider only sequences that can be described or generated from a finite string[15]. We define a *describable sequence* as any sequence $s$ where there exists an algorithm or computer $p$ such that for each element $e \in s$, $p$ generates $e$ in finite time/computation steps.[4]

With this set defined, we then want to be able to describe a particular sequence. To do this, we need to define a *universal description language*, or *udl*. A udl is defined as a 1:1 mapping between the set of finite strings from an alphabet $a$ and the set of describable sequences from a possibly different alphabet $b$ such that the mapping is describable. A mapping is describable if it can be generated by a universal computer given a finite description. To show that the set of describable sequences is unique for a given alphabet, we need only consult the definition of a universal computer, which says that it can do anything that any other computer can do[16].

Given all of this structure, we define a *model* as a probability distribution over the set of describable hypotheses such that there is no hypothesis where the probability is infinitesimal. We further define a *general model* as a model where no probabilities of individual hypotheses are equal to zero and a *special model* as a model where some hypothesis probabilities are equal to zero. As we will see, general models allow us the full power of general learning, whereas special models allow us to have smaller hypothesis spaces, such as in effectively all contemporary machine learning models.

In order to perform general induction, we need to be able to learn over the set of all *possible processes* rather than a subset that we believe to be most likely. If we were to restrict ourselves to a subset via a special model, we would be unable to learn that we are wrong about our original assumption, no matter how many observations we make. This is highly restrictive in general. Rather if we consider all possible processes, we can still represent our belief that some subset is more likely, but we are not restricted to considering only that subset should our beliefs turn out awry.[5] It is worth noting that IID processes may also be represented here via a process of marginalization to reflect independence[17].

We then define learning as the act of observing a subsequent element from a data sequence generated by a given process, removing the set of hypotheses that are inconsistent with the observation, and renormalizing the probability distribution by dividing each probability by the sum of the probabilities in the updated set. This is formally equivalent to a Bayesian learning process[18][19] and turns our model distributions into fully general learning algorithms[20].

We also want to be able to measure or evaluate the performance of a model. We can do this either in terms of how much probability it assigns to the true hypothesis or in terms of its predictive performance over unobserved data.

**Definition 1.** The *model-process correspondence* captures how closely the model distribution agrees with the process distribution. In the case of the fully deterministic processes we work with in this paper, this is simply the probability assigned by the model to the true data sequence.

**Definition 2.** The *model-process alignment* is the probability that a prediction made by the model will agree with data produced by the process. This is computed over all subsequent observations and all symbols in the alphabet using the model's probability distribution over hypotheses.

The idea behind correspondence is to capture the ability of the model to select the true hypothesis, whereas alignment captures the extent to which the model is able to predict future observations.

## Occam's razor

Given these definitions, we can now explore the nature of general learning. Our primary goal is to understand how induction proceeds despite underdetermination. Many hypotheses fit, few generalize.

---

[4]We limit ourselves to describable sequences because it is impossible to distinguish between describable and nondescribable concepts with finite data and it is more useful to assume describability.

[5]To simplify our study, we do not work directly with conditional processes nor with real valued data, though extensions can be made to our formalism to accommodate these and the results do not change categorically.

How do we choose? The umbrella principle we typically use for solving this general problem is Occam's razor or the principle of parsimony. It can be understood as a preference towards simpler hypotheses (or explanations) over more complex ones. When comparing hypotheses that fit the data equally, we often use Occam's razor arguments to motivate choosing the simpler hypothesis.

There are many formal approaches to these problems and they can be divided into two broad classes of Occam's razor arguments: *inductive Occam's razors* and *descriptive Occam's razors*. We will show that, while Occam's razors are useful in special cases, no Occam's razor argument is sufficient to ground induction in general, as the concept of simplicity is fundamentally contingent upon one's perspective and prior beliefs.

**Inductive Occam's razor**

Inductive Occam's razors include the simplicity principles utilized by VC theory, Rademacher complexity, AIC/BIC[21][22], and the Bayesian Occam's razor[23]. These all construct a concept of simplicity based around the amount of bias, or uncertainty in the model. While these are important concepts, it is well-known that this sense of simplicity is not capable of general model selection, as any increase in confidence (simplicity) has a corresponding decrease in confidence over other possible hypotheses[24]. While this is important, it diverges from the main thread of our argument and therefore further elaboration can be seen in the appendix.

**Descriptive Occam's razor**

Descriptive Occam's razor arguments are designed to mitigate the limitations of inductive Occam's razors. The idea is that some models are a priori more likely than others because they are fundamentally simpler. This is cached out by arguing that some models are shorter to describe, which is the focus of the theories of minimum description length[25], minimum message length[26], and Solomonoff induction[27][28]. To evaluate this idea, we have to introduce some concepts that allow us to think about the minimum description length of various models.[6]

We define the minimum description length of a hypothesis sequence $h$ in a given language $ul$ as the length of the shortest description string $d$ that causes language $ul$ to generate $h$. Formally, this is $\text{MDL}(ul, h) := min(len(d))$ s.t. $ul(d) = h$

Descriptive Occam's razor arguments in the algorithmic information theory tradition use these minimum description lengths to encode the relative simplicity of a hypothesis. Given these simplicity measures, we can formulate a probability distribution that favors simpler models over more complex ones. One way to do this is via the Solomonoff construction.

**Definition 3.** A *Solomonoff-model* is a general model derived from a universal language $ul$ via the Solomonoff construction, which gives the probability of a hypothesis $h$ as $p(h, ul) \propto |a|^{-\text{MDL}(ul,h)}$ where $|a|$ is the cardinality of the alphabet. For any universal language $ul$, we can refer to the canonical Solomonoff model for that language as $m_{ul}$.

Solomonoff uses his construction and the Bayesian learning process defined earlier to produce a general learning system that can, given enough data, learn to predict any sequential process in the sense that the limit of induction over infinite data is the true process[29][17]. This is true for any general model and can intuitively be understood as the increasing allocation of probability mass on the true hypothesis as additional observations continue to rule-out competing hypotheses. It can also be proven, via the algorithmic information theory invariance theorem, that switching to a different description language only affects the MDL of any given hypothesis by at most a constant factor dependent only upon the choice of the two languages[30]).

**Theorem 1** (Invariance). *For any two universal languages $ul1$ and $ul2$, and for any describable sequence $s$, the difference between the minimum description length of $s$ in $ul1$ and $s$ in $ul2$ is at most a constant $c$ dependent only upon the choice of $ul1$ and $ul2$.*

*Proof.* Universal languages are equivalent to universal computers. Universal computers ($uc \in \text{UC}$) have a property whereby one $uc$ can be used to simulate another $uc$ via only a finite description of the other $uc$. If we prepend this finite description to any description sequence, we can leverage the other $uc$ to generate our sequences. The only description length cost is this finite prepended string that is dependent only upon the choice of $uc$[31][32]. □

---
[6]The concepts and formalisms that we use are consistent with, but differ a bit from the associated concepts in algorithmic information theory. This is for clarity of presentation, given that a lot of the details in AIT are not directly relevant to this paper.

The effect of this invariance theorem on induction is to show that any differences in the prior probability distribution due to the choice of description language will eventually be overridden by Bayesian update after only a finite number of steps. This limits the effect of the choice of description language and gives us what can reasonably be called a "universal learning machine."

Typically the matter is left there, and further investigation goes into how to practically approximate such a universal learning machine or how to get a sense of how quickly the machine will learn[33]. But this is not the end of the story. There is an adversarial corollary to the invariance theorem which shows that general learning is not quite as it seems.

## Relativity

Though we can learn successfully from any choice of description language, the nature of any induction we perform may be dramatically affected by this choice. A universal simplicity measure that can justify induction should be able to prove that there exist some hypotheses that are universally more probable than others. We show that this is not possible in the following theorem.

**Theorem 2** (No universal probabilities). *For any finite set of hypotheses, the relative probability of these hypotheses in a Solomonoff prior is entirely dependent upon our choice of description language.*

*Proof.* This comes from the definition of a universal description language. Any describable mapping from description strings to hypothesis sequences is a valid language. It is straightforward to describe a mapping in which any finite set of strings maps arbitrarily to any finite set of sequences. □

More generally, the result we are looking for is simply that any describable mapping between these sets is an equally valid description language and therefore defines an a priori equivalent Solomonoff model. Since this includes reorderings of arbitrary finite sets, it is clear that there can be no universal probabilities or even relative probabilities assigned to any hypotheses.

While prior probabilities cannot have universal relative orderings, this might appear to be irrelevant, as some guarantee could conceivably emerge once we observe a sufficient amount of data. This is also not true. The following theorem shows that no finite amount of observed data can guarantee that the probability of any hypothesis is greater than the probability of any other.

**Theorem 3** (Choice of description language can overwhelm any finite data). *For any finite string of observed data and any finite set of describable hypotheses consistent with those data, the relative description lengths of the hypotheses can be arbitrarily modulated by the choice of udl.*

*Proof.* Similar to above. Choose a set of hypotheses consistent with the data. Apply theorem 2. □

It is also interesting to observe that a corollary of this theorem derives the generally assumed statistical fact that post-hoc inference is fallacious.

**Theorem 4** (Post-hoc fallacy). *For any observed data string $o$ and any describable prediction sequence $s$, it is always possible to choose a hypothesis $h$ such that $h = os$ ($o$ concatenated with $s$)*

*Proof.* $h$ is by definition describable. $o$ is describable because it is finite. We can construct a description for $os$ by describing the concatenation of $o$ and $s$. The existence of this description proves that it is in the universal hypothesis space. □

This shows that it is always possible to choose a model that makes a desired prediction (supports a hypothesis) and is consistent with known data. This is important not only for its role in helping us understand the post-hoc fallacy, but in how it offers a basis for relativity. It shows that we can always find a model that is consistent with known data and makes any prediction we choose. Similarly, we can show that, after any finite amount of observations, the set of possible unobserved sequences remains the same.

**Corollary 4.1** (Prior-posterior symmetry). *For any finite set of observations, the set of describable posteriors is the same as the set of describable priors.*

Given this result, our last hope for universal probability is to ask whether there are any a priori reasons to choose one description language over another. We can also show that there are none.

**Theorem 5** (No privileged language). *There does not exist a udl that is fundamentally simpler than any other in that it is shorter to describe or generates simpler descriptions overall.*

*Proof.* Solomonoff induction involves an assumed default description language $D$. The invariance theorem shows that any other UDL can be simulated by adding at most a finite constant. These alternative description languages $A$ therefore require overhead to simulate and are not as simple as $D$. By symmetry, if we choose any alternative UDL $a \in A$ as our default, $D$ becomes part of the alternatives set and is therefore more complex than $a$. For any choice of default description language, that description language becomes the most simple. □

**The relativity of induction**

But this does not ruin our ability to perform induction, it simply requires that we look at it from another perspective. As we saw in the descriptive Occam's razor section, it is possible to learn successfully given any choice of universal description language. We can now reconcile these results.

To understand how real world learning systems fit into this picture, it is helpful to introduce the concept of sequential *submodels*. For any general learning process, we can choose whether to view the system as a single learner receiving lots of data over long periods of time or as a sequence of sublearners connected via their prior and posterior distributions. In such a scheme, submodel $m_1$ starts with a random prior distribution, since we don't have any way of guessing this correctly. $m_1$ then makes some observations from the process and updates accordingly. The result is submodel $m_1^*$ and it most likely has higher correspondence with the process than submodel $m_1$. We can then define submodel $m_2$ such that its prior distribution is given by the posterior distribution $m_1^*$. $m_2$ goes through the same process as $m_1$ and results in $m_2^*$. We can iterate on this process arbitrarily many times until we reach $m_n$. This chain of belief transfer is equivalent to Bayesian inference. $m_n$ has a prior distribution determined by the learning that came before it, and we still do not know the alignment between this distribution and the process. We can, however, reasonably suspect that there is a decent level of alignment because the prior of $m_n$ is the result of a significant amount of learning that happened previously.[7] These submodels are far more like the models that we encounter everyday than the full general model is. This is because the prior beliefs of our learning systems do not come from a tabula rasa, but from previous learning. We can say that submodel $m_n$ operates relative to the chain of learners that came before it and which are responsible for its inductive bias.

Of course in the real world, such belief transfer is not limited to a long chain, but can be in the shape of an arbitrary causal graph[34]. Though we will not explore this formally here, it is important to be able to see how this applies. Because submodels exist only as part of a larger inductive system, and because they derive their properties from that system, scientists and engineers will benefit from thinking, not just in terms of local submodels, but about complete inductive *model systems*. With this view, we can look at particular components–e.g. scientists, machine learners, animals–and get a deeper understanding of their role in the greater distributed learning machine. Cognitive science has shown that cognition is often effectively understood in terms of large distributed systems with different dynamics emerging in particular subsystems[35], and it is the same way with learning and induction.

An important set of effects that emerge from such a system is a relative stability of the inductive bias. If $m_n$ is the result of a large amount of prior learning, then it has likely converged to capture most of the broader structure in the generating process. The remaining uncertainty lies mostly in the details. Due to the strong relations between prior distributions and simplicity concepts, our simplicity concepts also become quite stable. Stable and aligned simplicity concepts are useful tools and something that we can easily take for granted. Like the classic adage "the one thing a fish cannot see is the ocean", the contingency of the concept of similarity becomes invisible to downstream learners and it can easily be mistaken for a fundamental property of learning and induction.

We can alternatively think about the "simplicity" measure in Solomonoff induction as a perspective. Such a perspective captures the inductive biases that shape how we perceive a process and what appears to be simple vs. complex. The concepts of "perspective," "prior belief," "inductive bias," and "simplicity" are typically considered to be different phenomena, but in the general learning framework these distinctions disappear. Inductive biases are not just comprised by a choice of description language, but by a probability distribution over hypotheses. If we use the Solomonoff construction to define such a probability distribution, then the description language determines the distribution. However, the Solomonoff construction is just a canonical algorithm for defining general models, and it is far from the only one. To build a cleaner theoretical framework, we can eschew concepts of simplicity and description length in favor of describability and probability.

---
[7]We can even derive bounds for this[17]

## Applications

Humans are downstream learners and our beliefs are the products of a long chain of interactive entrainment processes. Therefore we should expect to have high model-process correspondence with the world and strong, useful concepts of simplicity and perspectives. We also create downstream learning systems in our peers, children, scientific theories, and machine learning systems.

### Culture and science

The human form is the result of an evolutionary process that has the same empirical structure as induction. Human culture is biased by the human form (neural structures, corporeal affordances, etc.), and shared experience across many humans who each update their beliefs through observation and share these discoveries with other humans[36][37][38]. The beliefs of individual humans are informed by their cultural context and serve as the cultural context for others[39][40]. All of this is progress from an original prior towards more effective posteriors. With each human's prior being informed by this immense amount of previous induction, it is no wonder that we are able to learn effortlessly. We benefit from the knowledge of our parents and peers and we pass what we have learned and integrated down to our children. It is a single distributed system that evolves and learns over time. As it is often said, "we stand on the shoulders of giants."

Science supervenes on top of this culture and has its own structures and processes for acquiring, sharing, and integrating knowledge. The design of scientific experiments is determined by prior beliefs and serves to provide maximum information update to the distributed scientific belief comprised by scientists, articles, and other information artifacts. Though describing this formally in terms of a general model is difficult, science gets all its belief from observation of the world, and therefore our theoretical framework directly applies. Because the scientific belief representation and update process is distributed, it only operates efficiently and effectively when these components are communicating and in sync. In order to achieve this, we benefit from sharing, not only the final designs and results of our experiments, but the motivations, assumptions, and prior beliefs that went into them. Explicating context and relativity is critical for effective scientific communication, as a statistical result taken without a context can be made to say anything. This information is all necessary to emulate an efficient Bayesian inductive process. If we don't coherently pass our learning forward in a way that it can be integrated into downstream priors, it is like it was never learned at all.

The relativity of description language also has a direct effect on the nature of science. If we write down a theory and it seems complex, we cannot disentangle the complexity of the theory from the framework we are using to describe it. Since there always exists an alternative description framework in which our theory appears very simple, the apparent simplicity vs. complexity of the theory is entirely dependent upon our perspective, and not on an inherent property of the theory. The upshot is that the merits of a scientific theory are relative to all of the observations and experience that went into forming them. A scientific theory must be judged based on the full scope of the process taken to reach it and not on its apparent simplicity or complexity.

### Machine learning

Machine learning models inherit an inductive bias that is the product of human experience, scientific knowledge, and iterated research. Ideally, when we design machine learning models, we are effectively representing all of our prior beliefs in the inductive bias of the model. However, most of our modeling paradigms are not explicitly Bayesian, so shaping the inductive bias becomes a bit more tricky. The inductive bias of a deep neural network emerges from the model architecture, weight initialization, optimization algorithm, training procedure, early stopping procedure, regularization strategy, and choice of loss function, and coordinating these to reflect your beliefs about the data-generating process is a complex craft. The present theory cannot say much directly about how to shape a deep neural network to emulate a general Bayesian submodel, but it can tell us that the more effectively you can represent our scientifically learned beliefs, the more likely your model is to have a high correspondence with the world. It can also help us to understand why, when this is successfully performed, deep learning works much better than other theories of learning would predict. Instead of evaluating models based on their effective complexity, we can think more about their expected correspondence. As a machine learning model's performance comes from a combination of the prior model-process correspondence and the amount of data observed, leveraging prior learning by the scientist and the research community will increase prior correspondence and require less data to achieve a given degree of predictive performance.

To improve our understanding of the importance of supplanting simplicity-based theory with correspondence principles, consider the case where we perform a regression either by fitting a line $y = Ax + b$ or by fitting a fixed-width parabola $y = x^2 + Ax + b$. Both have the exact same complexity on all of the traditional inductive complexity measures. However, for some data the line will be a much better model and the parabola will be a much better model for other data. Performance is modulated by the correspondence between the model and the world and the choice of which model to use is dependent upon the researcher's beliefs about the world. This is not just the case for toy examples; all models have their inductive biases, and the relationship between the inductive bias and the structure of the world is what dictates performance. Similarly, we can view the phenomenon of overfitting in light of the post-hoc fallacy theorem. Overfitting occurs when predictions are chosen with too much weight on whether hypotheses are consistent with so-far observed data and not enough emphasis on the prior beliefs. Though some view priors more or less as arbitrarily chosen black-boxes, they can be more accurately interpreted as representing learning that has occurred before the current ML model. In this way, the data scientist is a crucial part of the induction system and must be considered in the analysis.

Traditional learning theory doesn't capture the shape of the uncertainty in the model. If we represent the model as a distribution instead of a representation space, we can start to see that there is much more to a machine learning model than capacity-based theories would suggest[41]. A model is not equally likely to fit to all of the functions it can represent. This reflects a prior distribution over the space of representable hypotheses that emerges from the dynamics of the optimization process interacting with the model architecture and the data. If this prior distribution or inductive bias aligns highly with the data process, then it will take much less data to learn than if we were to start with a uniform distribution over the space of representable functions[18]. The role of a deep learning model is to represent the remaining uncertainty that a researcher has about a data process and to further lower it via observation. Seen from this perspective, the success of deep learning is not so mysterious.

## Conclusion

For too long have we rested on the concept of simplicity as our primary theoretical tool for understanding induction. While simplicity has its uses, it is not a fundamental principle, but a contingent property that emerges from stable belief structures. The key insight is to notice that no beliefs about the world may be derived from first principles. Rather we make observations, revise our understanding, and propagate it as evidence to be built upon by subsequent components of our great learning machine. Some belief structures may emerge as supported by great masses of evidence, in which case these structures may support equal weight of prediction and understanding. If we view learning as a process of convergence distributed across broad empirical systems, we can get a deeper understanding of how the properties of a learning system affect the performance of that system.

Human beliefs come not just from a crude trade between nature and nurture, but from a system of evolution and belief exchange. We inherit inductive biases genetically, in our environments, in our education, and via communication and we update these biases through observation, exploration, subsequent modification of our environment, and dissemination of our discoveries. Science operates in the same way, though it is scaffolded by additional formal processes and structured communication in order to facilitate the distributed instantiation and revision of theory. In machine learning, we take our prior beliefs and use them to construct learning systems that represent what we know and improve upon it via observation and update. As researchers and practitioners, we are as much a part of the machine learning systems as the models we study and deploy.

## Broader impact

As the prevalence and performance of deep learning methods has exploded in recent years, there has been a correspondingly widening gap between the accounts of our theory and the results of our experiments. Science works best when there is an interplay between theory and experiment–when theory suggests experiments that might yield interesting results and when experiments offer evidence to facilitate the revision of theory. The decay of this relationship in machine learning has meant that our experiments have become increasingly ad hoc and that they do not cohere into a broader understanding. While the prevalence of brilliant researchers in machine learning has allowed us to maintain steady and exciting progress, there remain massive fissures between the potential performance of learning machines and the performance of our current systems[42]. Natural language understanding is very far from producing systems that truly understand language, continual learning and meta-learning are in their infancy, and complex distributed machine learning systems that resemble humans embedded in a culture are nowhere to be found. Furthermore, our progress applies largely to isolated machine learning models, whereas it is known that a large part of "intelligent behavior" requires an integration between many kinds of data and knowledge to approach human performance by achieving AI completeness [43].

By broaching this gap between theory and practice with a new set of perspectives, we aim to facilitate new thinking about the nature of learning as well as novel approaches to the design of machine learning systems. If we can help build a theoretical framework that reflects what is actually going on in our machine learning systems–even if it is as messy as the processes themselves–then it will be easier for researchers to create new kinds of machine learners. For instance, recognition of the ways in which all learning is relativistic and distributed may directly inspire the development of technologies that improve our ability to share beliefs between machine learning subsystems as well as between humans and such systems. If these interfaces are more efficient, then the overall power of these learning systems will increase significantly and the space of problems that we can tackle in the field expands.

We are aware that some of the results in this article are rather informal, but our proximal purpose is not to supplant current theory with an alternative formal framework. Rather, we believe that the more critical project is to expand our current thinking about learning by illuminating phenomena and behaviors that govern the fundamentals of learning, but which have heretofore remained hidden from view. We hope that this article will serve as an impetus for further research–in both theory and practice–that will improve our understanding of our machines and ourselves as well as enhance our ability to build systems that lessen the problems of our age and further the evolution of humanity.

## References

[1] Behnam Neyshabur, Srinadh Bhojanapalli, David McAllester, and Nati Srebro. Exploring generalization in deep learning. In *Advances in Neural Information Processing Systems*, pages 5947–5956, 2017.

[2] Chiyuan Zhang, Samy Bengio, Moritz Hardt, Benjamin Recht, and Oriol Vinyals. Understanding deep learning requires rethinking generalization. *arXiv preprint arXiv:1611.03530*, 2016.

[3] Vladimir Vapnik. *The nature of statistical learning theory*. Springer science & business media, 2013.

[4] Peter L Bartlett and Shahar Mendelson. Rademacher and gaussian complexities: Risk bounds and structural results. *Journal of Machine Learning Research*, 3(Nov):463–482, 2002.

[5] Vladimir N Vapnik and A Ya Chervonenkis. On the uniform convergence of relative frequencies of events to their probabilities. In *Measures of complexity*, pages 11–30. Springer, 2015.

[6] Vaishnavh Nagarajan and J Zico Kolter. Uniform convergence may be unable to explain generalization in deep learning. In *Advances in Neural Information Processing Systems*, pages 11611–11622, 2019.

[7] Behnam Neyshabur, Ryota Tomioka, and Nathan Srebro. In search of the real inductive bias: On the role of implicit regularization in deep learning. *arXiv preprint arXiv:1412.6614*, 2014.

[8] E Sober. *Ockam's Razors: A User's Manual*. New York, NY: Cambridge University Press, 2015.

[9] Alan Baker. Simplicity. In Edward N. Zalta, editor, *The Stanford Encyclopedia of Philosophy*. Metaphysics Research Lab, Stanford University, winter 2016 edition, 2016.

[10] Posterior Analytics. translated by grg mure in the basic works of aristotle, edited by richard mckeon, 1941.

[11] Marvin Minsky and Seymour A Papert. *Perceptrons: An introduction to computational geometry*. MIT press, 2017.

[12] Georges Rey. The analytic/synthetic distinction. In Edward N. Zalta, editor, *The Stanford Encyclopedia of Philosophy*. Metaphysics Research Lab, Stanford University, fall 2018 edition, 2018.

[13] Corinna Cortes and Vladimir Vapnik. Support-vector networks. *Machine learning*, 20(3):273–297, 1995.

[14] Claude E Shannon. A mathematical theory of communication. *Bell system technical journal*, 27(3):379–423, 1948.

[15] Peter D Grünwald, Paul MB Vitányi, et al. Algorithmic information theory. *Handbook of the Philosophy of Information*, pages 281–320, 2008.

[16] Alan Mathison Turing. On computable numbers, with an application to the entscheidungsproblem. *J. of Math*, 58(345-363):5, 1936.

[17] Ming Li, Paul Vitányi, et al. *An introduction to Kolmogorov complexity and its applications*, volume 3. Springer, 2008.

[18] David Haussler, Michael Kearns, and Robert E Schapire. Bounds on the sample complexity of bayesian learning using information theory and the vc dimension. *Machine learning*, 14(1):83–113, 1994.

[19] Jürgen Schmidhuber. Algorithmic theories of everything. *arXiv preprint quant-ph/0011122*, 2000.

[20] M. Hutter, S. Legg, and P. M.B. Vitanyi. Algorithmic probability. *Scholarpedia*, 2(8):2572, 2007. revision #151509.

[21] Hirotogu Akaike. Information theory and an extension of the maximum likelihood principle. In *Selected papers of hirotugu akaike*, pages 199–213. Springer, 1998.

[22] Gideon Schwarz et al. Estimating the dimension of a model. *The annals of statistics*, 6(2):461–464, 1978.

[23] David JC MacKay and David JC Mac Kay. *Information theory, inference and learning algorithms*. Cambridge university press, 2003.

[24] Cullen Schaffer. A conservation law for generalization performance. In *Machine Learning Proceedings 1994*, pages 259–265. Elsevier, 1994.

[25] Peter D Grünwald and Abhijit Grunwald. *The minimum description length principle*. MIT press, 2007.

[26] Chris S Wallace and David M Boulton. An information measure for classification. *The Computer Journal*, 11(2):185–194, 1968.

[27] Ray J Solomonoff. A formal theory of inductive inference. part i. *Information and control*, 7(1):1–22, 1964.

[28] Ray J Solomonoff. A formal theory of inductive inference. part ii. *Information and control*, 7(2):224–254, 1964.

[29] Samuel Rathmanner and Marcus Hutter. A philosophical treatise of universal induction. *Entropy*, 13(6):1076–1136, 2011.

[30] Andrei N Kolmogorov. Three approaches to the quantitative definition of information'. *Problems of information transmission*, 1(1):1–7, 1965.

[31] Shane Legg. Solomonoff induction. Technical report, Technical Report 30, Centre for Discrete Mathematics and Theoretical . . . , 1997.

[32] Gregory J Chaitin. On the simplicity and speed of programs for computing infinite sets of natural numbers. *Journal of the ACM (JACM)*, 16(3):407–422, 1969.

[33] Marcus Hutter. *Universal artificial intelligence: Sequential decisions based on algorithmic probability*. Springer Science & Business Media, 2004.

[34] Judea Pearl. *Probabilistic reasoning in intelligent systems: networks of plausible inference*. Elsevier, 2014.

[35] Edwin Hutchins. *Cognition in the Wild*. MIT press, 1995.

[36] James Jerome Gibson. *The senses considered as perceptual systems.* Houghton Mifflin, 1966.

[37] Richard A Watson and Eörs Szathmáry. How can evolution learn? *Trends in ecology & evolution*, 31(2):147–157, 2016.

[38] Thomas L Griffiths, Michael L Kalish, and Stephan Lewandowsky. Theoretical and empirical evidence for the impact of inductive biases on cultural evolution. *Philosophical Transactions of the Royal Society B: Biological Sciences*, 363(1509):3503–3514, 2008.

[39] Michael Cole. *Cultural psychology: A once and future discipline.* Harvard University Press, 1998.

[40] Harry C Triandis. *The analysis of subjective culture.* Wiley-Interscience, 1972.

[41] Shai Shalev-Shwartz and Shai Ben-David. *Understanding machine learning: From theory to algorithms.* Cambridge university press, 2014.

[42] Jeff Clune. How meta-learning could help us accomplish our grandest ai ambitions, and early, exotic steps in that direction. NeurIPS, 2019.

[43] John C Mallery. Thinking about foreign policy: Finding an appropriate role for artificially intelligent computers. In *Master's thesis, MIT Political Science Department*. Citeseer, 1988.

[44] Ron Kohavi, David H Wolpert, et al. Bias plus variance decomposition for zero-one loss functions. In *ICML*, volume 96, pages 275–83, 1996.

[45] Pedro Domingos. A unified bias-variance decomposition. In *Proceedings of 17th International Conference on Machine Learning*, pages 231–238, 2000.

[46] David H Wolpert. The supervised learning no-free-lunch theorems. In *Soft computing and industry*, pages 25–42. Springer, 2002.

[47] Tom M Mitchell. *The need for biases in learning generalizations*. Department of Computer Science, Laboratory for Computer Science Research . . . , 1980.

# Appendix

**Inductive Occam's razor**

Inductive Occam's razor arguments construct a concept of simplicity (we will call this the inductive complexity) that depends on the amount of "bias"[44][45] in the model. Classically, this has been thought of in terms of the set of hypotheses that can be represented by the model. For example, the VC theory framework operationalizes model complexity in terms of the VC dimension, which uses the set of binary output functions that can be represented by a given class of models. Other conceptions of inductive Occam's razor are less explicit about the set construction and find other ways to capture the expressivity of the model class or hypothesis space. These include Rademacher complexity, AIC/BIC, and the Bayesian Occam's razor.

In general, we can think about the inductive complexity of a model in terms of a quantification of uncertainty about the true hypothesis concept. This requires us to construct a probability distribution over hypotheses. In the case of strict set-theoretic or non-probabilistic concepts of inductive complexity such as VC dimension, this is represented by simply assigning a probability of zero to all hypotheses that cannot be represented by the model and an equal probability to the rest. In the more general case, we simply have a probability distribution over the space of possible hypotheses. Given such a distribution, we can think about the inductive complexity of a model as the entropy of this distribution.[8]

This conception intuitively makes sense because a model with less uncertainty is more confident and will require less data to learn if the true concept is consistent with the model's belief. For instance, if we are highly confident that the true hypothesis is a linear function, it will require less data for us to learn the true linear function than if we are only confident that it is some sort of polynomial. This is compatible with existing measures of model complexity and extends them. Of course the opposite is true as well. If we have a low-complexity–high confidence–model and are incorrect in our beliefs, it will require far more data to learn the true concept than if we were less confident to begin with.

This trade-off is at the center of our argument for the inadequacy of inductive complexity as a general principle of learning. It turns out that any increase in confidence (simplicity) of our model has a corresponding decrease in confidence over other possible hypotheses. This is a good thing if we are correct, but hurts performance if we are not. In the fully agnostic setting where we do not know anything a priori about the true hypothesis, by increasing or decreasing our model's confidence, we are simply trading off risk, not expected reward. Similar arguments have been demonstrated many times and are well-known[46][24][47].

The distinction between models in terms of different amounts of bias is useful, but it does not give us the tools to solve the underdetermination of induction problem. Inductive Occam's razor allows us to determine which models are more complex, not which models are more likely.

---

[8]To properly formalize this, the general case requires a modified concept to allow us to distinguish between various infinite entropies.